\documentclass[letterpaper, 10 pt, conference]{ieeeconf}  %

\IEEEoverridecommandlockouts
\usepackage[pdftex]{graphicx}
\graphicspath{{figures/}}
\usepackage{hyperref}
\hypersetup{
    colorlinks=true,
    linkcolor=black,
    citecolor=black,
    filecolor=black,
    urlcolor=black,
}
\usepackage[T1]{fontenc}
\usepackage[utf8]{inputenc}
\usepackage{csquotes}
\usepackage[english]{babel}
\usepackage[export]{adjustbox}
\usepackage{caption}
\usepackage{subcaption}
\usepackage{amsmath}
\usepackage{amssymb}
\usepackage{tabularx}
\usepackage{multirow}
\usepackage[table, dvipsnames]{xcolor}
\usepackage[colorinlistoftodos]{todonotes}
\usepackage{lipsum}
\usepackage[capitalize]{cleveref}
\usepackage{balance}
\usepackage{placeins}%

\usepackage[style=ieee,
            doi=false,
            url=false,
            mincitenames=1,
            maxcitenames=1,
            minbibnames=6,
            maxbibnames=6,
            backend=biber]{biblatex}  %
\title{%
\textsc{MAP-Former}: Multi-Agent-Pair Gaussian Joint Prediction
}

\author{
    Marlon Steiner$^{1}$,
    Marvin Klemp$^{1}$,
    and Christoph Stiller$^{1}$%
    \thanks{
        $^{1}$Marlon Steiner, Marvin Klemp and Christoph Stiller are with the Institute of Measurement and Control Systems, Karlsruhe Institute of Technology (KIT),
        Karlsruhe, Germany
        {\tt\small \{marlon.steiner,marvin.klemp,stiller\}@kit.edu}
    }%
}

\usepackage{textcomp}
\usepackage[absolute,showboxes]{textpos}
\TPMargin{5pt}

\newcommand{\copyrightstatement}{
    \begin{textblock}{0.82}(0.09,0.93)
         \noindent{\footnotesize{\copyright 2024 IEEE.
         Personal use of this material is permitted.
         Permission from IEEE must be obtained for all other uses, in any current or future media, including reprinting/republishing this material for advertising or promotional purposes, creating new collective works, for resale or redistribution to servers or lists, or reuse of any copyrighted component of this work in other works.

         \vspace{2mm}
         \noindent
         Accepted for publication in Proceedings of the IEEE Intelligent Vehicles Symposium (IV), Jeju Island - Korea, 2-5 June 2024.}}
    \end{textblock}
}

\AtBeginBibliography{\small}

\begin{document}
\maketitle

\copyrightstatement

\thispagestyle{empty}
\pagestyle{empty}

\begin{abstract}

There is a gap in risk assessment of trajectories between the trajectory information coming from a traffic motion prediction module and what is actually needed.
Closing this gap necessitates advancements in prediction beyond current practices.
Existing prediction models yield joint predictions of agents' future trajectories with uncertainty weights or marginal Gaussian probability density functions (PDFs) for single agents.
Although, these methods achieve high accurate trajectory predictions, they only provide little or no information about the dependencies of interacting agents.
Since traffic is a process of highly interdependent agents, whose actions directly influence their mutual behavior, the existing methods are not sufficient to reliably assess the risk of future trajectories.
This paper addresses that gap by introducing a novel approach to motion prediction, focusing on predicting agent-pair covariance matrices in a ``scene-centric'' manner, which can then be used to model Gaussian joint PDFs for all agent-pairs in a scene.
We propose a model capable of predicting those agent-pair covariance matrices, leveraging an enhanced awareness of interactions.
Utilizing the prediction results of our model, this work forms the foundation for comprehensive risk assessment with statistically based methods for analyzing agents' relations by their joint PDFs.

\end{abstract}
\section{Introduction}
\label{introduction}
\begin{figure}[!h]
    \begin{subfigure}{0.49\columnwidth}
    \includegraphics[width=\linewidth, trim=15cm 4cm 8cm 2cm, clip]{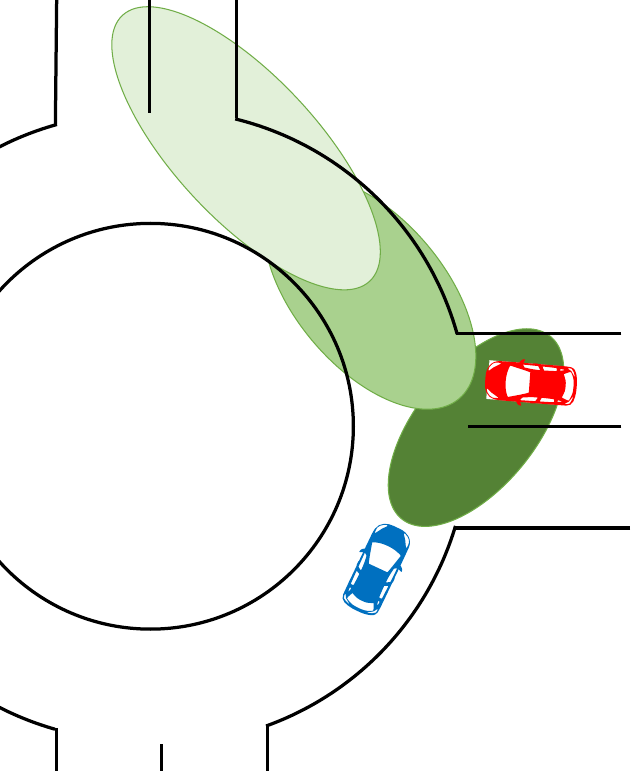}
    \caption{Mode 1: green}
    \label{fig:gaussianPrediction_a}
    \end{subfigure}
    \hfill
    \begin{subfigure}{0.49\columnwidth}
    \includegraphics[width=\linewidth, trim=15cm 4cm 8cm 2cm, clip]{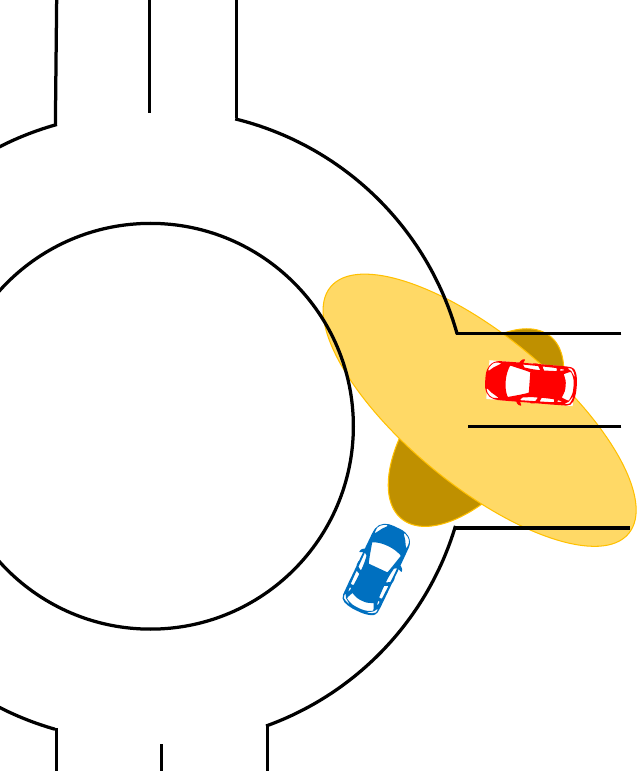}
    \caption{Mode $n$: yellow}
    \label{fig:gaussianPrediction_b}
    \end{subfigure}

    \caption{Representation of the prediction model's output:
    One Gaussian joint PDF for every time step and every mode (\ref*{fig:gaussianPrediction_a}, \ref*{fig:gaussianPrediction_b}) of an agent-pair based on the predicted covariance matrices.
    The different shadings of the ellipses (joint PDFs) represent the consecutive time steps.
    Due to visualization reasons, 2d ellipses are used instead of 4d Gaussian PDFs as the model actually predicts.
    Combining the modes with uncertainty weights results in a Gaussian mixture PDF.}
    \label{fig:gaussianPrediction}
\end{figure}

Motion planning has a strong dependency to motion prediction.
Hence, every planned trajectory is influenced by a guess of how the current scene will evolve in the next seconds.
The prediction of future trajectories is a non-trivial task:
Every action an agent takes, influences its neighboring and scene-related agents, and thus propagates information through the scene.
Accordingly, there is an interdependency between actions of all agents in a scene.
For the risk assessment of trajectories it is therefore crucial to represent the statistical dependencies between agents in their predicted trajectories.
We determine these dependencies by developing a model, which is able to predict agent-pair covariance matrices for the $x$ and $y$ coordinates of both vehicles (\cref*{fig:gaussianPrediction}).

\begin{figure}[h]
    \includegraphics[width=\linewidth, trim=2cm 5cm 2cm 2cm, clip]{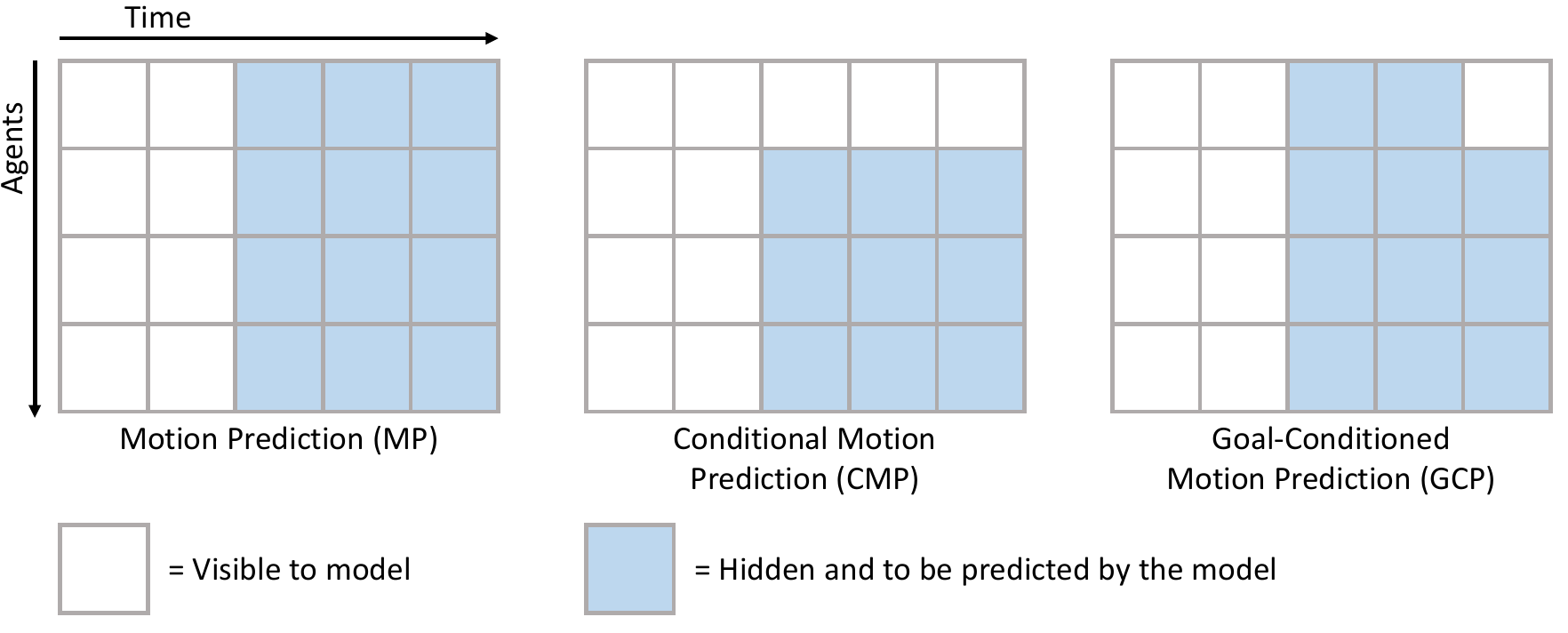}
    \caption{\label{fig:MP} Different tasks in motion prediction.
    Here the first two columns represent the past and the last three the future.}
\end{figure}

This work is focused on the prediction fundamentals, required for a statistical risk assessment method of trajectories.
The motion prediction task can be distinguished between motion prediction (MP), conditional motion prediction (CMP) and goal-conditioned motion prediction (GCP).
\cite*{ngiam2022scenetransformer} provides a very descriptive figure to portray the differences of the three motion prediction tasks.
A similar illustration is provided in \cref*{fig:MP}.
The figure shows the motion prediction tasks as three grids of white and blue pixels with the time axis on the horizontal and the agent axis on the vertical.
While the white pixels symbolize that the corresponding value can be accessed by the model, the blue pixels are hidden to the model and must be predicted.
In the classic MP, the past trajectories of all agents are known by the model and the future must be predicted.
This can be done jointly respectively in a ``scene-centric'' manner or marginal for only one agent.
The former one is currently experiencing increasing interest, especially because of its higher significance compared to just predicting marginal trajectories.

In CMP, the whole future of the ego-agent is known, and all other agents must be predicted.
While CMP can be a valid assumption for marginal trajectory planning of an ego-agent, we implement MP, since the goal is to predict scene-centric and not focusing on a single agent.
Also, this method might be deprecated, since this does not reflect the reality where driving requires multidirectional responses.

Finally, GCP also defines one agent as the ego-agent but in contrast to CMP it only provides information about its last step.
We think, that the assumption made in GCP is only valid for a comparatively long prediction horizon, otherwise the provided goal position is less an information of destination but more an information of how the scene is evolving.
E.g, considering a prediction horizon of $5\,\mathrm{s}$ in an interactive scene would implicitly provide the prediction module the information of how the interaction is solved.
This gives the model more than just routing information, which are in reality not accessible and thus limiting the responsiveness of real traffic.

We choose the rounD dataset \cite*{krajewski2020rounD} because roundabouts represent highly interactive traffic scenarios, which is desirable for our purpose.
The dataset contains tracks of $13\,746$ agents (including cars, vans, trucks, buses, pedestrians, bicycles, motorcycles) and has a total length of six hours.
Further, the rounD dataset provides high definition (HD) maps with nodes and edges representing properties of the road.
We have also deliberately avoided the usage of standard datasets like \textsc{Waymo Open Motion Dataset} \cite*{ettinger2021waymoOpenMotion}, \textsc{Argoverse} \cite*{chang2019argoverse} or \textsc{nuScenes} \cite*{caesar2020nuScenes} because we explicitly do not focus on the best prediction results regarding a leaderboard but provide a social aware method to predict agent-pair covariance matrices.

To the best of our knowledge we are the first predicting covariance matrices for agent-pairs.
Compared to just predicting these for single agents, this provides further statistical information about the relations of agents.
To summarize, the main contributions of this work are:
\begin{enumerate}
    \item A novel model architecture of combining spatial, interaction and temporal information with a GNN and a factorized Transformer for motion prediction.
    \item Predicting covariance matrices for agent-pairs in a scene-centric manner, so that multivariate Gaussian joint PDFs can be constructed for all agent-pairs in a scene.
    \item A covariance matrix formulation, that can be used in machine leaning, while guarantying its mathematical properties and a corresponding multivariate Gaussian negative log likelihood loss formulation.
    \item Proposal of using predicted covariance matrices as a foundation for statistical interactivity and risk analysis.
\end{enumerate}

\section{Related Work}
\label{sec:related_work}
The huge amount of publications in the field of motion prediction reflects its enormous relevance in research and application.
While our long-term goal is the usage of prediction models in motion planning and especially in risk assessment of trajectories, we still provide a quick overview about recent research in MP.

\subsubsection*{\textbf{Marginal Prediction}}
MP started with predicting marginal future trajectories, i.e. trajectories for single agents in a scene.
Different approaches have been used to build the prediction models:
Most recent approaches use graph-based methods with GNNs, like \textsc{VectorNet} \cite*{gao2020VectorNet}, \textsc{LaneGCN} \cite*{liang2020LaneGCN} and \textsc{LaneRCNN} \cite*{zeng2021lanercnn} and  \textsc{Multipath++} \cite*{varadarajan2022multipath++} or \textsc{Transformer}-based \cite*{vaswani2017Transformer} architectures, like \textsc{Wayformer} \cite*{nayakanti2023wayformer}.
While \cite*{gao2020VectorNet}, \cite*{liang2020LaneGCN} and \cite*{zeng2021lanercnn} state results on marginal prediction metrics, the authors claim, that joint prediction is possible with their models.
Earlier approaches made predictions based on CNNs, e.g. \textsc{Multipath} \cite*{chai2019multipath} and \cite*{cui2019multimodal}.
All those prediction models consider multi-modality and therefore output multiple modes of trajectories.

\subsubsection*{\textbf{Joint Prediction}}
In joint prediction, Transformer-based models like \textsc{Scene Transformer} \cite*{ngiam2022scenetransformer} and \mbox{\textsc{AgentFormer} \cite*{yuan2021agentformer}} dominate over other architectures regarding the prediction metrics.
Those models predict future trajectories in a scene-centric manner, so that they output multi-modal futures for all agents in a scene.
An example for GNN-based joint prediction is the \textsc{JFP} model \cite*{luo2023jfp}.
Instead of outputting a joint prediction as a single distribution like in \cite*{ngiam2022scenetransformer}, agents get modeled pairwise in \cite*{luo2023jfp} and based on that, the whole joint distribution is build up sequentially.

Due to its superior performance, Transformer-based architectures gain increasing interest and have been widely and successfully used for MP tasks.
Nevertheless, there is still ongoing research in GNN-based architectures due to their ability of modeling the environmental and social components.

The latest research direction regarding MP was proposed by \cite*{seff2023motionlm}.
They make use of language models (LMs) for predicting joint future trajectories with their model \textsc{MotionLM}.
Therefore, multi-agent rollouts over discrete motion tokens are leveraged, capturing the joint distribution over multimodal futures.

\subsubsection*{\textbf{Gaussian Prediction}}
Next to predicting just coordinate-based trajectories in a marginal or joint manner, research also focuses on predicting PDFs (mostly Gaussians).
This section provides literature, which is most related to our work.
Examples for predicting Gaussian mixture (GM) PDFs are \textsc{WIMP} \cite*{khandelwal2020whatif}, \textsc{SAMMP} \cite*{mercat2020JointForecasting}, \textsc{CBP} \cite*{tolstaya2021} and \textsc{Wayformer} \cite*{nayakanti2023wayformer}.
Like most of the approaches in MP, they also perform multi-modal predictions.
Those models share the fact that they predict Gaussian PDFs for single agents and therefore, provide uncertainty information only for a respective agent.
Both, the \textsc{WIMP} model \cite*{khandelwal2020whatif} and the \textsc{CBP} model \cite*{tolstaya2021} are GNN-based approaches and perform multi-modal prediction as a CMP task with social, environmental and temporal information.
The \textsc{SAMMP} model \cite*{mercat2020JointForecasting} is solely based on vehicle position tracks, and utilizes multi-head attention mechanisms in combination with LSTMs for its prediction.

\subsubsection*{\textbf{Prediction in Roundabout Scenarios}}
As we evaluate our method on roundabouts in the rounD dataset, this section lists papers which also perform prediction on roundabouts.
\cite*{janjovs2021self}, \cite*{westny2023evaluation} and \cite*{daoud2023prediction} also train their prediction models on the rounD dataset.
Another dataset providing roundabout scenarios is the \mbox{INTERACTION} dataset \cite*{zhan2019interaction}.
Examples for prediction models evaluated on this dataset are \cite*{scibior2021imagining} and \cite*{grimm2023holistic}.
Except of \cite*{grimm2023holistic}, all presented papers train and compete their models on marginal prediction.
\section{Method}
\begin{figure*}[t!]
    \includegraphics[width=\textwidth]{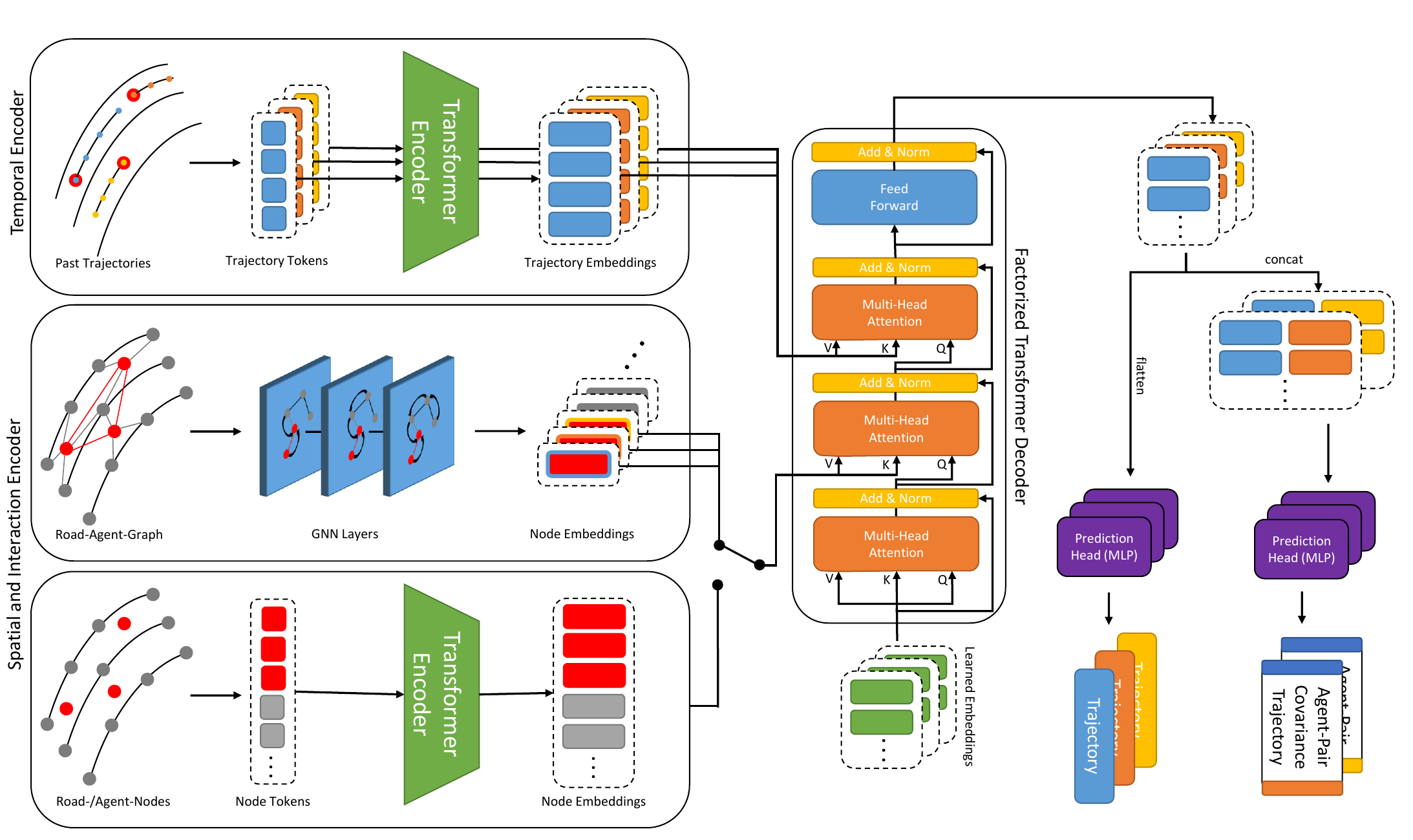}
    \caption{\label{fig:architecture} Network architecture of our motion prediction model.
    We use a \textit{TEnc} (top left) and the different models can switch between either a GNN-based \textit{SaIEnc} (middle left), a Transformer-based \textit{SaIEnc} (bottom left) or no \textit{SaIEnc}.
    The red points in the encoders represent the agents.
    The colors blue, orange and yellow associated with the tokens, embeddings and trajectories represent the corresponding agents.}
\end{figure*}

In comparison to the presented Gaussian prediction models, which model multiple modes of each agent as a GM PDF, we are predicting multi-modal PDFs for all relevant agent-pairs, simultaneously.
This makes our approach more suitable for assessing risk of trajectories, because the agent-pair PDFs provide more relational information about an agent-pairing.
Also, the ego-agent could, e.g. analyze PDFs of other agent-pairs for its own motion planning.
Additionally, our approach differs from \cite*{khandelwal2020whatif} and \cite*{tolstaya2021} that we do scene-centric joint prediction instead of marginal conditional prediction.

Our multi-agent-pair Transformer model (\textsc{MAP-Former}) is composed of four main modules (see \cref*{fig:architecture}):
(1) The \textit{Temporal Encoder} uses a Transformer encoder to embed the past trajectories into high dimensional space.
(2) The second module is the \textit{Spatial and Interaction Encoder}, which uses a graph as input and extracts its information with either a GNN-based architecture or a Transformer-based architecture.
(3) The \textit{Factorized Transformer Decoder} applies cross-attention to the outputs of (1) and (2) with learned embeddings for the future trajectories.
\mbox{(4) The} last module produces the predictions and consists of two sets of $n$ linear prediction heads:
The first set predicts multiple modes of future trajectories for every agent in a scene.
While the second set predicts the parameters of a covariance matrix for every agent-pair corresponding to their trajectories.
Both is predicted simultaneously in a single feed-forward pass.
The agent-pair covariance matrix prediction is the core of our work, since it allows us to model Gaussian joint PDFs for all agent-pairs in a scene.

\subsection{Temporal Encoder (TEnc)}
The \textit{TEnc} processes information about the past trajectories of all agents (red points in \cref*{fig:architecture}) in a scene.
For every agent $A$, the corresponding past trajectory is visualized in a different color, which matches to the color of the embeddings in the rest of the figure.
Taking every agent's past trajectory point as a separate token and stacking the tokens of different agents into different vectors, produces the input to the encoder.
As the encoder, a Transformer encoder is used.
To process the tokens in the encoder, every token gets converted into a unique embedding, so that the past trajectory embedding is of the shape $[A, n_{\mathrm{pastTimeSteps}}, d_{\mathrm{model}}]$.
Here $d_{\mathrm{model}}$ represents the embedding dimension used over all modules in the model.
The encoder applies self-attention to each embedding within a trajectory and again outputs embeddings for every agent and every time step.

\subsection{Spatial and Interaction Encoder (SaIEnc)}
To include structural and relational information of a scene, we use a second encoder which either is a GNN-based \textit{SaIEnc} or a Transformer-based \textit{SaIEnc}.

\subsubsection*{\textbf{GNN-based SaIEnc}}
Since, GNNs have proven strong performance (e.g. \cite*{gao2020VectorNet}, \cite*{liang2020LaneGCN}) in learning structural relations, we implement a GNN-based \textit{SaIEnc} to capture further contextual information.
For this encoder, we provide a directed road-agent-graph consisting of nodes and edges as input.
Nodes represent agents (red points in \cref*{fig:architecture}) and also structural elements of the road (gray points).
Edges serve as a connection of contextually coherent nodes, so that connected nodes can aggregate information from each other.
The road-graphs are build up from HD maps.

The agent-graph is implemented as a fully connected graph (see \cref*{fig:architecture}), where every agent can aggregate information from all other agents.
To enable the aggregation of map information, every agent is further connected to all road-graph nodes within a radius of $r=5\,\mathrm{m}$ to its center.
All nodes and edges are defined by an individual feature vector, characterizing their properties or in case of an edge, the type of connection.

The GNN layers are implemented based on the GIN \cite*{xu2018gin} architecture, where the feature vector $h_v^{(k)}$ of node $v$ in layer $k$ is calculated as follows:
\begin{equation*}
    h_v^{(k)} = \mathrm{MLP}^{(k)}\left(h_v^{(k-1)} + \sum_{u\in\mathcal{N}(v)}h_u^{(k-1)} \right).
\end{equation*}
The sum is formed over the set $\mathcal{N}(v)$ of neighboring nodes $u$ of the node of interest $v$.
It can be guaranteed that different nodes with different neighborhoods are always mapped as different embeddings, since the MLP and the sum operator, which operates on multisets, are injective.
A GNN layer can be seen as a message passing algorithm, where every node receives messages from its neighbors and aggregates them into a new feature vector.
For every GNN layer $k$ added to the network, the receptive field consequently increases by one ``hop''.

The original GIN implementation is not meant to process edge features.
Therefore, we use the PyTorch \mbox{Geometric \cite*{fey2019pyg}} version of the GIN network (GINE), which can also handle edge features in the message passing.

The output of the GNN are embeddings for every node in the graph of shape $[n_\mathrm{nodes}, d_\mathrm{model}]$, with encoded information about their $k$-hop neighbors.
\cref*{fig:architecture} visualizes the agent embeddings bordered with the same colors as in the \textit{TEnc}.
The node embeddings of the road-graph (gray points) are discarded.

\subsubsection*{\textbf{Transformer-based SaIEnc}}
As an alternative to the previous encoder, we use a Transformer-based encoder to capture the contextual information of the scene \cite*{wagner2023redmotion}.
Here, the nodes of the road-agent-graph are directly used as input.
The Transformer encoder applies self-attention between the node embeddings, to extract the structural and relational information between the nodes.

\subsection{Factorized Transformer Decoder}
The \textit{Factorized Transformer Decoder} aggregates the information of both, the \textit{TEnc} and the \textit{SaIEnc}, and outputs the final embeddings per agent.
We use learned embeddings (green in \cref*{fig:architecture}) of the shape $[A, n_{\mathrm{futureTimeSteps}}, d_{\mathrm{model}}]$ as another input to the decoder, which attach to the encoder embeddings.

First, self-attention is applied among the learned embeddings to identify the relevant relations between the time step embeddings per agent.
The output of the self-attention is then used as the query in the first cross-attention block.
Here, the node embeddings from the \textit{SaIEnc} serve as agent specific values and keys.
Thus, the learned embeddings can attend to the relevant structural and social information.
In the next step, the output is again used as the query for the second cross-attention block.
The values and keys are taken from the \textit{TEnc}, so that the learned embeddings get updated by attending to the past trajectory embeddings.
Analogous to \cite*{vaswani2017Transformer} the whole decoder block is repeated $N$ times.

\subsection{Multihead Agent-Pair Prediction}
This module takes the generated embeddings of the \textit{Factorized Transformer Decoder} as its input.
For the trajectory prediction this information is directly fed into $n=6$ MLP heads to predict multiple modes of future trajectories.
Whereas, for predicting the agent-pair covariance matrices, we first form agent-pairs.
Therefore, we choose one agent per scene serving as the ``ego-vehicle'', whose embedding is then concatenated with all other agent embeddings.
E.g. in \cref*{fig:architecture}, the blue embedding (blue agent) is concatenated with the orange embedding respectively the yellow embedding.
The concatenation results in an embedding of shape \mbox{$[A-1, n_{\mathrm{futureTimeSteps}}, 2\cdot d_{\mathrm{model}}]$}.
Our architecture also allows to extend the concatenation for all existing agent-pairs in a scene.
In the next step, the agent-pair embeddings are fed into $n$ MLP prediction heads, resulting in agent-pair covariance matrices $\Sigma$ corresponding to the predicted trajectories.

To guaranty, that the predicted covariance matrices fulfill the requirements on symmetry and positive-definiteness, we utilize the properties of the Cholesky decomposition:
A symmetric and positive-definite matrix can be decomposed with a lower triangle matrix $\mathbf{L}$ and a positive-definite diagonal matrix $\mathbf{D}$ to:
\begin{equation*}
    \Sigma= \mathbf{L}\mathbf{D}\mathbf{L}^\mathrm{T}.
\end{equation*}
Adapted to our case with four coordinates as random variables, the matrices $\mathbf{L}$ and $\mathbf{D}$ can be constructed as following:
\begin{equation*}
    \mathbf{L} =
    \begin{bmatrix}
        1 & 0 & 0 & 0 \\
        a & 1 & 0 & 0 \\
        b & c & 1 & 0 \\
        d & e & f & 1 \\
    \end{bmatrix}
        ,\ \mathbf{D} =
    \begin{bmatrix}
        \hat{\sigma}_{x_1}^2 & 0 & 0 & 0 \\
        0 & \hat{\sigma}_{y_1}^2 & 0 & 0 \\
        0 & 0 & \hat{\sigma}_{x_2}^2 & 0 \\
        0 & 0 & 0 & \hat{\sigma}_{y_2}^2 \\
    \end{bmatrix}.
\end{equation*}

With this formulation of the covariance matrix $\Sigma$, we need to predict ten parameters:
$\hat{\sigma}_{x_1}\,\hat{\sigma}_{y_1}\,\hat{\sigma}_{x_2}\,\hat{\sigma}_{y_2}\in\mathbb{R}^+$ and the parameters $a,\,b,\,c,\,d,\,e,\,f\in\mathbb{R}$ of the matrix $\mathbf{L}$.
The standard deviations $\sigma$ of the covariance matrix cannot directly be chosen by the $\hat{\sigma}$ in $\mathbf{D}$, but these parameters significantly influence the standard deviations.
Standard deviations take on values \mbox{$\sigma>0$} per definition.
This property is modeled by using a \textit{Softplus} activation function.
Afterwards, we shift the output by adding a fixed bias, under which the standard deviation is physically not reasonable.

\subsection{Multivariate Gaussian Negative Log Likelihood Loss}
By taking the predicted coordinates $\mu$ and the covariance matrix $\Sigma$ of a single time step and considering the four coordinates $x_1,\,y_1,\,x_2,\,y_2$ as random variables $\mathrm{X}$, the density function of a multivariate Gaussian PDF can be constructed:
\begin{equation*}
    f_\mathbf{X}(x_1,y_1,x_2,y_2) = \frac{\mathrm{exp}\!\left(-\frac{1}{2}(\mathbf{x}-\mu)^\mathrm{T}\Sigma^{-1}(\mathbf{x}-\mu)\right)}{\sqrt{(2\pi)^k\,\mathrm{det}(\Sigma)}}.
\end{equation*}
Here $k$ represents the dimension of $\mathrm{X}$.
As a modification of the Gaussian negative log likelihood loss (GNLL), we form the multivariate case of this loss (MGNNL):
\begin{align*}
    &loss = -\mathrm{log}\left(f_\mathbf{X}\right) \\
        &= \mathrm{log}\!\left(\sqrt{(2\pi)^k\,\mathrm{det}(\Sigma)} \right) + \left(\frac{1}{2}(\mathbf{x}-\mu)^\mathrm{T}\Sigma^{-1}(\mathbf{x}-\mu)\right).
\end{align*}

The GNLL loss is derived from the assumption, that the target values are normally distributed around the predicted values.
The loss measures how well the predicted PDF explains the predicted values.
It encourages the model to not only make accurate predictions but also provide appropriate uncertainty estimates and thus fit a reasonable distribution.

This loss formulation is the core of our work, since the goal is to receive a covariance matrix for all future time steps of every agent-pair.
Due to the properties of the Cholesky decomposition, this matrix is always symmetric and positive-definite.
Thus, the inversion of the covariance matrix $\Sigma^{-1}$ for the loss calculation can be guaranteed.

\section{Evaluation}
\label{sec:evaluation}

As mentioned before, we train our models on the rounD dataset.
Since we perform agent-pair prediction, we skip all frames in which only a single agent occurs.
The maximum number of agents recorded in the dataset for a single frame is 25.
Hence, our model predicts 2-25 agents jointly, depending on the frame.
We predict trajectory points in a frequency of $5\,\mathrm{Hz}$ and provide $1\,\mathrm{s}$ of history to the model.

For the evaluation of our prediction results, we use the standard metrics but in an extended way, so that they are capable of capturing the scene-centric joint prediction:
\begin{itemize}
    \item Minimum Scene-Centric Average Displacement Error (minSADE) in meters:
    The ADE is calculated as the average Euclidean L2 distance between all points of a predicted trajectory and the corresponding points of the ground truth.
    The SADE is therefore the mean over all ADEs of a specific mode.
    Therefore, the min refers to the mode that provides the minimum SADE.
    \item Minimum Scene-Centric Final Displacement Error (minSFDE) in meters:
    The FDE is the L2 distance between the endpoint of a predicted trajectory and its ground truth.
    Analogous to the SADE, the SFDE is the mean over all FDEs of a specific mode.
    Consequently, the minSFDE takes the minimum over all predicted modes.
    \item Scene-Centric Miss Rate (SMR):
    A prediction counts as a ``miss'', when the FDE is larger than $2\,\mathrm{m}$.
    In our case, the SMR is calculated as the number of scenarios where the FDE of at least one agent of the mode with the lowest SFDE is larger than $2\,\mathrm{m}$, divided by the total number of scenarios.
\end{itemize}

\begin{table*}[t]
    \vspace*{1mm}
    \center
    \caption{Results on rounD dataset for $n=6$ prediction heads (or modes) and a prediction horizon of $t=3\,\mathrm{s}$ and $t=5\,\mathrm{s}$ with ``scene-centric'' (S) metrics.
             The models with gray background are only evaluated on marginal prediction (not scene-centric). %
             The best joint prediction results are highlighted in bold and the best marginal prediction results are underlined.}
    \begin{tabularx}{11.15cm}{p{4cm} | c c | c c | c c}
        \hline\noalign{\vskip 1mm}
        \multirow{2}{4cm}{\centering Method} & \multicolumn{2}{c|}{minSADE $\downarrow$} & \multicolumn{2}{c|}{minSFDE $\downarrow$} & \multicolumn{2}{c}{SMR $\downarrow$} \\
         & $3\,\mathrm{s}$ & $5\,\mathrm{s}$ & $3\,\mathrm{s}$ & $5\,\mathrm{s}$ & $3\,\mathrm{s}$ & $5\,\mathrm{s}$ \\\noalign{\vskip 1mm}
        \hline\noalign{\vskip 1mm}
        \rowcolor{gray!20} \textsc{SSP-ASP} \cite*{janjovs2021self} & \underline{0.17} & 1.25 ($6\,\mathrm{s}$) & \underline{0.74} & 4.61 ($6\,\mathrm{s}$) & - & - \\[1mm]
        \rowcolor{gray!20} \textsc{N-ODE2} \cite*{westny2023evaluation} & - & \underline{0.98} & - & \underline{3.09} & - & \underline{0.35} \\[1mm]
        \rowcolor{gray!20} Extended \textsc{DGNN} \cite*{daoud2023prediction} & 1.68 & - & 1.69 & - & - & - \\\noalign{\vskip 1mm}
        \textsc{CNN} Baseline \cite*{nikhil2018convolutional} & 1.46 & 3.57 & 4.30 & 10.29 & 1.00 & 1.00 \\[1mm]
        \hline\noalign{\vskip 1mm}
        \textsc{MAP-Former} (Baseline) (Ours) & 0.71 & 1.49 & 1.84 & 4.03 & 0.90 & 0.97 \\[1mm]
        \textsc{MAP-GraphFormer} (Ours) & 0.60 & 1.30 & 1.59 & 3.48 & 0.83 & 0.96 \\[1mm]
        \textsc{MAP-Former} (full) (Ours) & \textbf{0.52} & \textbf{1.20} & \textbf{1.38} & \textbf{3.22} & \textbf{0.75} & \textbf{0.95} \\\noalign{\vskip 1mm}
        \hline
    \end{tabularx}
    \label{tab:results}
\end{table*}

\cref*{tab:results} shows the performance of different models on the rounD dataset.
We propose three different models resulting from the \textsc{MAP-Former} architecture (\cref*{fig:architecture}):
First, the \textsc{MAP-Former} (Baseline), which only uses the \textit{TEnc} as an encoder.
Second, the \textsc{MAP-GraphFormer}, which uses the \textit{TEnc} and the GNN-based \textit{SaIEnc}.
And third, the \textsc{MAP-Former} (full), which uses the \textit{TEnc} and the Transformer-based \textit{SaIEnc}.

For comparison, we implement a simple joint prediction \textsc{CNN} Baseline \cite*{nikhil2018convolutional}.
To the best of our knowledge, we are the first performing joint prediction on the rounD dataset.
Therefore, the results of three marginal prediction models (\cite*{janjovs2021self}, \cite*{westny2023evaluation}, \cite*{daoud2023prediction}) are provided (gray background in \cref*{tab:results}).

Adding information about the scene structure to our model, as done with the \textit{SaIEnc}, improves the prediction performance.
While the \textsc{MAP-GraphFormer} represents an improvement over the \textsc{MAP-Former} (Baseline), the \textsc{MAP-Former} (full) outperforms the given joint prediction models in all metrics.
The results of the marginal models and the joint models are not directly comparable.
Anyway, the \textsc{MAP-Former} (full) competes with the best marginal prediction model (\textsc{N-ODE2} \cite*{westny2023evaluation}) in the longer prediction horizon.
Only in the short prediction horizon, the \textsc{MAP-Former} (full) is outperformed by the \textsc{SSP-ASP} \cite*{janjovs2021self} model.
Notably, the \textsc{SSP-ASP} and the \textsc{N-ODE2} models get a history of $3\,\mathrm{s}$ instead of $1\,\mathrm{s}$ like our models.

\begin{figure*}[t!]
    \includegraphics[width=\textwidth, trim=6.4cm 3.3cm 6.9cm 3cm, clip]{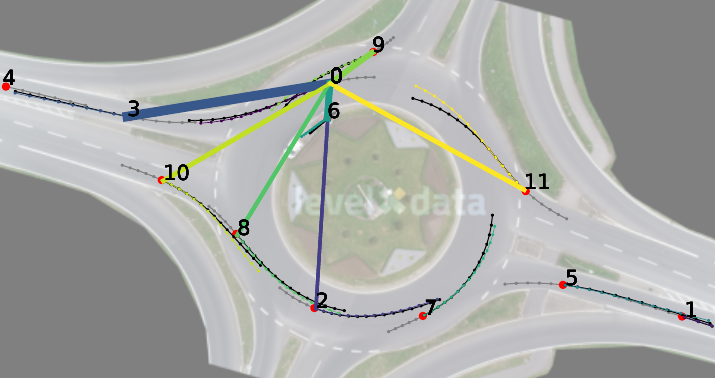}
    \caption{\label{fig:results} Visual prediction results:
    The figure shows a scene from rounD \cite*{krajewski2020rounD} with twelve agents (red points).
    For every agent the figure provides its past trajectory (gray), its ground truth (black) and its predicted trajectory for $t=3\,\mathrm{s}$ (colored).
    The lines, connecting the agent-pairs, represent the upper diagonal blocks of the predicted covariance matrices and therefore describe the dependencies between agent-pairs.
    }
\end{figure*}

In the following, we shortly present the results for the agent-pair covariance matrix prediction.
A covariance matrix ($4\times4$) can be seen as a $2\times2$ block matrix with four $2\times2$ blocks.
The blocks on the main diagonal represent the marginal covariance matrices of the predicted trajectory points of the respective agents within an agent-pair.
While the block on the upper diagonal (identical to the lower diagonal block) represents the covariance matrix between the agents.
For a first proof of concept, we sum up the absolute values of the upper diagonal blocks of the predicted agent-pairs' covariance matrices.
These sums are shown in \cref*{fig:results} as lines with variable thickness connecting the agents.
Where the thickness represents a measure of the dependency respectively the interactivity between agent-pairs.
The predictions originate from our \textsc{MAP-Former} (full) model.

In this example we build up agent-pairs based on agent 0 and visualized possibly interesting dependencies in \cref*{fig:results}.
Agent 6 inside the roundabout is a human, maintaining the planted area.
We can see that agent 0 has the lowest dependency with agent 2, which is reasonable because agent 2 is spatially far away from agent 0 and there are also agents between them.
Also, agent 8 has a low dependency with agent 0, since agent 8 has already entered the roundabout and agent 0 is about to exit.
Agent 10 and 11 have a higher and quite similar dependency with agent 0.
Given that agent 0 is about to exit the roundabout, agent 10 and 11 are in a similar situation to agent 0.
The highest dependencies to agent 0 have -- in ascending order -- agents 6, 9 and 3.
Agent 9 is spatially close to agent 0 and therefore, a high dependency is reasonable.
It is possibly unsure, if agent 6 intends to cross the road, and it is additionally spatially close related to agent 0, so a high dependency is also reasonable.
Agent 3 is spatially further away from agent 0, but it can be considered as the leading vehicle and thus has a high dependency with agent 0.
A comprehensive statistical analysis of the agent-pair Gaussian PDFs will be part of a follow-up work, so we will not go into more detail about the agent-pair correlation analysis.

\section{Conclusions and Future Work}
\label{sec:conclusion}
In this work we presented a novel view on motion prediction in the context of motion planning and risk assessment.
We propose to predict statistical information between agent-pairs.
Therefore, we developed a multi-agent-pair prediction model, capable of predicting not only coordinates of a trajectory but predicting joint covariance matrices.
These can be used for modeling agent-pair Gaussian PDFs and calculate dependencies between them.
While Gaussian PDFs for single agents only allow a statement about the uncertainties of each agent's coordinates, our method also provides information about their dependencies.
Based on this prediction approach, a follow-up paper will comprehensively analyze agent interactivity and risk utilizing the probabilistic joint PDFs generated by the predicted covariance matrices.
\section*{Acknowledgements}
This work is accomplished within the project \mbox{``AUTOtech.agil''} (FKZ 01IS22088) and the financial support from the German Federal Ministry of Education and Research (BMBF) is acknowledged.
 
\balance
\printbibliography

\end{document}